\documentclass{article}

\usepackage{microtype}
\usepackage{graphicx}
\usepackage{subfigure}
\usepackage{booktabs}
\usepackage{verbatim}
\usepackage{pdfpages}
\usepackage{amsmath}
\usepackage{tikz}
\usepackage{pdfpages}
\usepackage{array}
\usepackage{arydshln}
\usepackage{diagbox}
\usepackage{multirow, makecell}
\usepackage{rotating}
\usepackage{tabularx}
\usepackage{multirow}
\usepackage{amssymb}
\usepackage[mathscr]{eucal}
\usepackage{adjustbox}
\usepackage{comment}
\usepackage[hyphens]{url}
\usepackage{bibentry}

\usepackage{comment}
\usepackage{color}
\usepackage{regexpatch} 
\usepackage[colorinlistoftodos,prependcaption]{todonotes}
\makeatletter
\xpatchcmd{\@todo}{\setkeys{todonotes}{#1}}{\setkeys{todonotes}{inline,#1}}{}{}
\makeatother

\newcolumntype{R}[2]{%
    >{\adjustbox{angle=#1,lap=\width-(#2)}\bgroup}%
    l%
    <{\egroup}%
}

\usepackage{hyperref}

\newcommand\mysign{\smash{{\scalebox{0.8}{\textless}}}}
\newcommand{\s}{\stackrel{\text{\fontsize{5}{0}\selectfont MFR}}{\mysign}}

\usepackage[accepted]{icml2019}

\newcolumntype{Y}{>{\centering\arraybackslash}X}
\newcolumntype{M}{>{\raggedleft\arraybackslash}X}
\newcolumntype{C}[1]{>{\centering}p{#1}}
\newcolumntype{O}[1]{>{\raggedright}p{#1}}
\newcolumntype{Q}[1]{>{\raggedleft}p{#1}}

\icmltitlerunning{Context-Aware Zero-Shot Learning for Object Recognition}

\begin{document}

\twocolumn[
\icmltitle{Context-Aware Zero-Shot Learning for Object Recognition}



\icmlsetsymbol{equal}{*}

\begin{icmlauthorlist}
\icmlauthor{\'Eloi Zablocki}{equal,su}
\icmlauthor{Patrick Bordes}{equal,su}
\icmlauthor{Benjamin Piwowarski}{su}
\icmlauthor{Laure Soulier}{su}
\icmlauthor{Patrick Gallinari}{su,cr}
\end{icmlauthorlist}

\icmlaffiliation{su}{Sorbonne Universit\'e, CNRS, Laboratoire d'Informatique de Paris 6, LIP6, F-75005 Paris, France}
\icmlaffiliation{cr}{Criteo AI Lab, Paris} 

\icmlcorrespondingauthor{\'Eloi Zablocki}{eloi.zablocki@lip6.fr}

\icmlkeywords{Zero-Shot Learning, Visual Context, Object Recognition, Visual Genome}

\vskip 0.3in
]



\printAffiliationsAndNotice{\icmlEqualContribution} 

\begin{abstract}
Zero-Shot Learning (ZSL) aims at classifying unlabeled objects by leveraging auxiliary knowledge, such as semantic representations. 
A limitation of previous approaches is that only intrinsic properties of objects, e.g.\ their visual appearance, are taken into account while their context, e.g.\ the surrounding objects in the image, is ignored. 
Following the intuitive principle that objects tend to be found in certain contexts but not others, we propose a new and challenging approach, \textit{context-aware ZSL}, that leverages semantic representations in a new way to model the conditional likelihood of an object to appear in a given context. 
Finally, through extensive experiments conducted on Visual Genome, we show that contextual information can substantially improve the standard ZSL approach and is robust to unbalanced classes. 
\end{abstract}

\vspace{-.5cm}
\section{Introduction}
Traditional Computer Vision models, such as Convolutional Neural Networks (CNNs) \cite{Lecun98gradient-basedlearning}, are designed to classify images into a set of predefined classes.
Their performances have kept improving in the last decade, namely on object recognition benchmarks such as ImageNet \cite{DBLP:conf/cvpr/DengDSLL009}, where state-of-the-art models \cite{DBLP:journals/corr/ZophVSL17,DBLP:journals/corr/abs-1802-01548} have outmatched humans.
However, training such models requires hundreds of manually-labeled instances for each class, which is a tedious and costly acquisition process.
Moreover, these models cannot replicate humans' capacity to generalize and to recognize objects they have never seen before. 
As a response to these limitations, Zero-Shot Learning (ZSL) has emerged as an important research field in the last decade \cite{DBLP:conf/cvpr/FarhadiEHF09,DBLP:conf/eccv/MensinkVPC12,DBLP:journals/corr/FuYHXG15,DBLP:conf/cvpr/KodirovXG17}.
In the object recognition field, ZSL aims at labeling an instance of a class for which no supervised data is available, by using knowledge acquired from another disjoint set of classes, for which corresponding visual instances are provided. In the literature, these sets of classes are respectively called \textit{target} and \textit{source} domains --- terms borrowed from the transfer learning community.
Generalization from the source to the target domain is achieved using auxiliary knowledge that semantically relates classes of both domains, e.g. attributes or textual representations of the class labels.


Previous ZSL approaches only focus on intrinsic properties of objects, e.g. their visual appearance, by the means of handcrafted features --- e.g. shape, texture, or color --- \cite{DBLP:journals/pami/LampertNH14} or distributed representations learned from text corpora \cite{DBLP:journals/pami/AkataPHS16,DBLP:conf/cvpr/LongLSSDH17}. 
The underlying hypothesis is that the identification of entities of the target domain is made possible thanks to the implicit \textit{principle of compositionality} (a.k.a.\ Frege's principle \cite{frege}) --- an object is formed by the composition of its attributes and characteristics --- and the fact that other entities of the source domain share the same attributes.
For example, if textual resources state that an apple is round and that it can be red or green, this knowledge can be used to identify apples in images because these characteristics (`round`, `red`) could be shared by classes of the source domain (e.g, `round` like a ball, `red` like a strawberry\dots).


We believe that visual context, i.e. the other entities surrounding an object, also explains human's ability to recognize an object that has never been seen before.
This assumption relies on the fact that scenes are \emph{compositional} in the sense that they are formed by the composition of objects they contain.
Some works in Computer Vision have exploited visual context to refine the predictions of classification \cite{Mensink2014COSTACS} or detection \cite{DBLP:conf/cvpr/BellZBG16} 
models.
To the best of our knowledge, context has not been exploited in ZSL because, for obvious reasons, it is impossible
to directly estimate the likelihood of a context for objects from the target domain --- from visual data only.
However, textual resources can be used to provide insights on the possible visual context in which an object is expected to appear.
To illustrate this, knowing from language that an apple is likely to be found hanging on a tree or in the hand of someone eating it, can be very helpful to identify apples in images. 
In this paper, our goal is to leverage visual context as an additional source of knowledge for ZSL, by exploiting the distributed word representations \cite{mikolov2013} of the object class labels.
More precisely, we adopt a probabilistic framework in which the probability to recognize a given object is split into three components:
(1) a \textit{visual component} based on its visual appearance (which can be derived from any traditional ZSL approach),
(2) a \textit{contextual component} exploiting its visual context, 
and (3) a \textit{prior component}, which estimates the frequency of objects in the dataset.
As a complementary contribution, we show that separating prior information in a dedicated component, along with simple yet effective sampling strategies, leads to a more interpretable model, able to deal with imbalanced datasets.
Finally, as traditional ZSL datasets lack contextual information, we design a new dedicated setup based on the richly annotated Visual Genome dataset \cite{visualgenome}. We conduct extensive experiments to thoroughly study the impact of contextual information.



\vspace{-0.2cm}
\section{Related work}
\label{related_work}

\paragraph{Zero-shot learning}
\label{zero_shot}


While state-of-the-art image classification models \cite{DBLP:journals/corr/ZophVSL17,DBLP:journals/corr/abs-1802-01548} restrict their predictions to a finite set of predefined classes, ZSL bypasses this important limitation by transferring knowledge acquired from seen classes (\textit{source domain}) to unseen classes (\textit{target domain)}.
Generalization is made possible through the medium of a common semantic space where all classes from both source and target domains are represented by vectors called \textit{semantic representations}.

Historically, the first semantic representations that were used were handcrafted attributes
\cite{DBLP:conf/cvpr/FarhadiEHF09,DBLP:conf/cvpr/ParikhG11,DBLP:conf/eccv/MensinkVPC12,DBLP:journals/pami/LampertNH14}. In these works, the attributes of a given image are determined and the class with the most similar attributes is predicted. 
Most methods represent class labels with binary vectors of visual features (e.g, 'IsBlack','HasClaws') \cite{DBLP:conf/cvpr/LampertNH09,DBLP:conf/cvpr/LiuKS11,DBLP:journals/pami/FuHXG14,DBLP:journals/pami/LampertNH14}.
However, attribute-based methods do not scale efficiently since the attribute ontology is often domain-specific and has to be built manually.
To cope with this limitation, more recent ZSL works rely on distributed semantic representations learned from textual datasets such as Wikipedia, using Distributional Semantic Models \cite{mikolov2013,glove,elmo}.
These models are based on the \textit{distributional hypothesis} \cite{harris1954distributional}, which states that textual items with similar contexts in text corpora tend to have similar meanings. This is of particular interest in ZSL: all object classes (from both source and target domains) are embedded into the same continuous vector space based on their textual context, which is a rich source of semantic information.
Some models directly aggregate textual representations of class labels and the predictions of a CNN \cite{DBLP:journals/corr/NorouziMBSSFCD13}, 
whereas others learn a cross-modal mapping between image representations (given by a CNN) and pre-learned semantic embeddings \cite{DBLP:conf/cvpr/AkataRWLS15,DBLP:conf/eccv/BucherHJ16}.
At inference, the predicted class of a given image is the nearest neighbor in the semantic embedding space. 
The cross-modal mapping is linear in most of ZSL works \cite{DBLP:conf/nips/PalatucciPHM09,DBLP:conf/icml/Romera-ParedesT15,DBLP:journals/pami/AkataPHS16,DBLP:conf/cvpr/QiaoLSH16}; this is the case in the present paper.
Among these works, the DeViSE model \cite{devise} uses a max-margin ranking objective to learn a cross-modal projection and fine-tune the lower layers of the CNN. 
Several models have built upon DeViSE with approaches that learn non-linear mappings between the visual and textual modalities \cite{DBLP:conf/iccv/BaSFS15,DBLP:conf/cvpr/XianA0N0S16}, or by using a common multimodal space to embed both images and object classes \cite{DBLP:conf/cvpr/FuXKG15,DBLP:conf/cvpr/LongLSSDH17}. 
In this paper, we extend DeViSE in two directions: by additionally leveraging visual context, and by reformulating it as a probabilistic model that allows coping with an imbalanced class distribution.

\begin{figure*}[t]
\centering
    \includegraphics[width=0.9\linewidth]{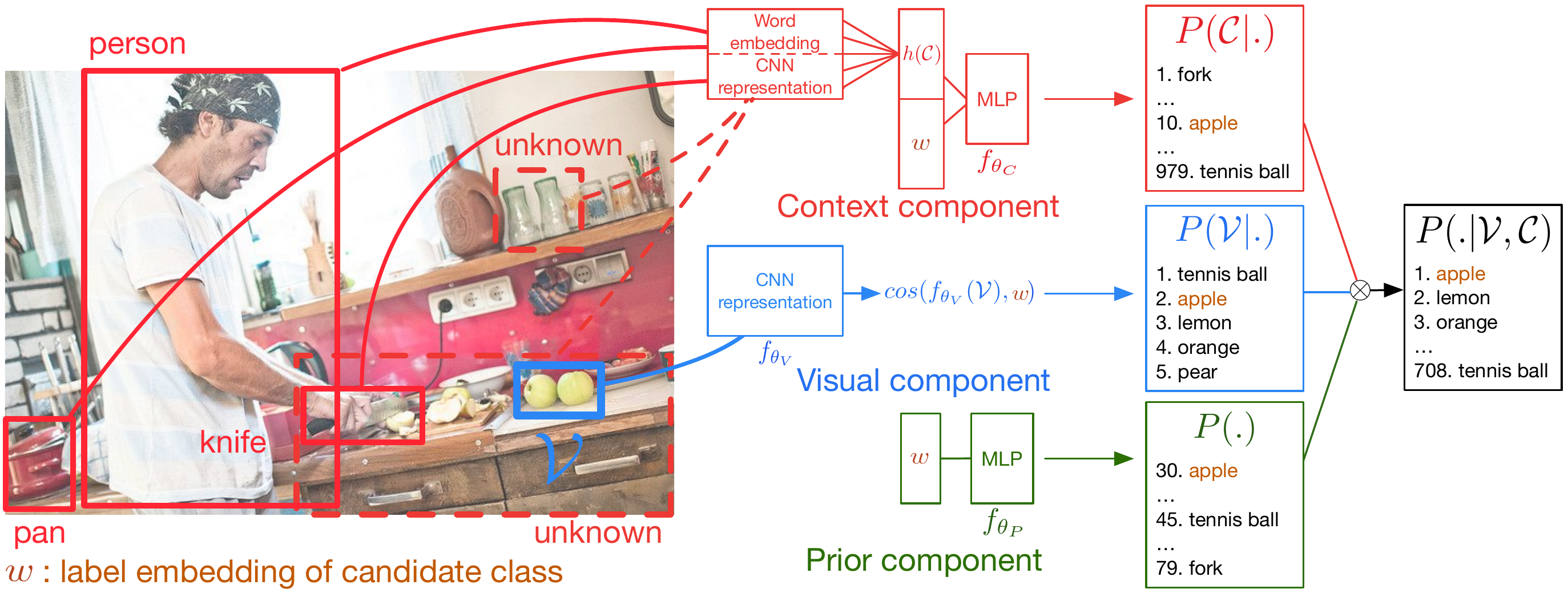}
    \vspace{-0.4cm}
    \caption{The goal is to find the class (in the target domain) of the object contained within the blue image region $\mathcal{V}$. Its context is formed of labeled objects from the source domain (red plain boxes) and of unlabeled object from the target domain (red dashed boxes). 
    }
    \vspace{-0.4cm}
    \label{model_overview}
\end{figure*}


\vspace{-0.4cm}
\paragraph{Visual context} 
\label{context_object}

The intuitive principle that \textit{some objects tend to be found in some contexts but not others}, is
at the core of many works. In NLP, visual context of objects can be used to build efficient word representations \cite{zablockiaaai2018}.
In Computer Vision, it can be used to refine detection \cite{DBLP:journals/pami/ChenSDHHY15,DBLP:journals/ijon/ChuC18} or segmentation \cite{contextsegm2018} tasks.

Visual context can either be \textit{low-level} (i.e.\ raw image pixels) or \textit{high-level} (i.e.\ labeled objects). When visual context is exploited in the form of \textit{low-level} information \cite{DBLP:journals/ijcv/Torralba03,DBLP:journals/ijcv/WolfB06,DBLP:journals/cacm/TorralbaMF10}, it often consists of global image features. 
For instance, in \cite{DBLP:conf/cvpr/HeZC04}, a Conditional Random Field is trained at combining low-level image features to assign to each pixel a class. 
In \textit{high-level} approaches, the referential meaning of the context objects (i.e. class labels) is used.
For example, \citet{DBLP:conf/iccv/RabinovichVGWB07} show that high-level context can be used at the post-processing level to reduce the ambiguities of a pre-learned object classification model, by leveraging co-occurrence patterns between objects that are computed from the training set.
Moreover, \citet{the_role_of_context_selection} study the role of context to classify objects: they investigate the importance of contextual indicators, such as object co-occurrence, relative scale and spatial relationships, and find that contextual information can sometimes be more informative than direct visual cues from objects.
Spatial relations between objects can also be used in addition to co-occurrences, as in \cite{DBLP:conf/cvpr/GalleguillosRB08,DBLP:conf/cvpr/ChenLFG18}.  
In \cite{DBLP:journals/corr/BengioDEILRSS13}, co-occurrences are computed using external information collected from web documents. The model classifies all objects jointly; it gives an inference method enabling a balance between an image coherence term (given by an image classifier) and a semantic term (given by a co-occurrence matrix). However, the approach is fully supervised, and this setting cannot be applied to ZSL.
The {context-aware zero-shot learning} task is related to the graph generation tasks \cite{DBLP:conf/cvpr/ZellersYTC18,DBLP:conf/eccv/YangLLBP18} and visual relationship detection \cite{DBLP:conf/eccv/LuKBL16}.

In conclusion, while many works 
in NLP and Computer Vision show the importance of visual context, its use in ZSL remains a challenge, that we propose to tackle in this paper.

\vspace{-0.2cm}
\section{Context-aware Zero-Shot Learning}
\label{model}

Let $\mathcal{O}$ be the set of all object classes, divided in classes from the \textit{source domain} $\mathcal{S}$ and classes from the \textit{target domain} $\mathcal{T}$. 
The goal of our approach --- \textit{context-aware ZSL} --- is to determine the class $i \in \mathcal{T}$ of an object contained in an image $I$, given its visual appearance $\mathcal{V}$ and its visual context $\mathcal{C}$. 
The {image} $I$ is annotated with bounding boxes, each containing an object.
Given the zone $\mathcal{V}$, the context $\mathcal{C}$ consists of the surrounding objects in the image. Their classes can either belong to the source domain ($\mathcal{C}\cap \mathcal{S}$) or to the target domain ($\mathcal{C}\cap \mathcal{T}$). 
Note that the class of an object of $\mathcal{C} \cap \mathcal{T}$ is not accessible in ZSL, only its visual appearance is.

\vspace{-0.2cm}
\subsection{Model overview}

We tackle this task by modeling the conditional probability $P(i|\mathcal{V},\mathcal{C})$ of a class $i$ given both the visual appearance $\mathcal{V}$ and the visual context $\mathcal{C}$ of the object of interest.
Given the absence of data in the target domain, we need to limit the complexity of the model, for generalizability's purpose. 
Accordingly, we suppose that $\mathcal{V}$ and $\mathcal{C}$ are conditionally independent given the class $i$ --- we show in the experiments (section~\ref{results}) that this hypothesis is acceptable.
This hypothesis leads to the following expression:
\vspace{-0.1cm}
\begin{equation}
P(i|\mathcal{V},\mathcal{C})
\propto
P(\mathcal{V} | i)
P(\mathcal{C}| i )
P(i)
\label{p_global}
\end{equation}
where each conditional probability expresses the probability of either the visual appearance $\mathcal{V}$ or the context $\mathcal{C}$ given class $i$, and $P(i)$ denotes the prior distribution of the dataset. Each term of this equation is modeled separately.

The intuition behind our approach is illustrated in Figure~\ref{model_overview}, where the blue box contains the object of interest. Here, the class is \textit{apple}, which belongs to the target domain $\mathcal{T}$.
The visual component, which focuses on the zone $\mathcal{V}$, recognizes a \textit{tennis ball} due to its yellow and round appearance; \textit{apple} is ranked second. The prior component indicates that \textit{apple} is slightly more frequent than \textit{tennis ball}, but the frequency discrepancy may not be high enough to change the prediction of the visual component. In that case, the context component is discriminant: it ranks objects that are likely to be found in a kitchen, and reveals that an \textit{apple} is far more likely to be found than a \textit{tennis ball} in this context.

Precisely modeling $P(\mathcal{C}|.)$, $P(\mathcal{V}|.)$ and $P(.)$ is challenging due to the ZSL setting. Indeed, these distributions cannot be computed for classes of the target domain because of the absence of corresponding training data.
Thus, to transfer the knowledge acquired from the source domain to the target domain, we use a common semantic space, namely \textit{Word2Vec} \cite{mikolov2013}, where source and target class labels are embedded as vectors of $\mathbb{R}^d$, with $d$ the dimension of the space.
It is worth noting that we propose to separately learn the prior class distribution $P(.)$ with a ranking loss (in section \ref{learning}). This allows dealing with imbalanced datasets, in contrast to ZSL models like DeViSE \cite{devise}. This intuition is experimentally validated in section \ref{scenarios}. 

\vspace{-0.2cm}
\subsection{Description of the model's components}
\label{modeling}

Due to both the ZSL setting and the variety of possible context and/or visual appearance of objects, it is not possible to estimate directly the different probabilities of equation \ref{p_global}.
Hence, in what follows, we estimate quantities related to $P(\mathcal{C}|.)$, $P(\mathcal{V}|.)$ and $P(.)$
using parametric energy functions \cite{lecun2006tutorial}.
These quantities are learned separately, as described in section~\ref{learning}. 
Finally, we explain how we combine them to produce the global probability $P(.|\mathcal{C}, \mathcal{V})$ in section~\ref{inference}.

\vspace{-0.2cm}
\paragraph{Visual component}
The visual component models $P(\mathcal{V}|i)$ by computing the compatibility between the visual appearance $\mathcal{V}$ of the object of interest, and the semantic representation $w_i$ of the class $i$.

Following previous ZSL works based on cross-modal projections \cite{devise,DBLP:conf/eccv/BansalSSCD18}, we introduce $f_{\theta_V}$, a parametric function mapping an image to the semantic space: $f_{\theta_V}(\mathcal{V})=W_V.\texttt{CNN}(\mathcal{V})+b_V \in \mathbb{R}^d$
where $\texttt{CNN}(\mathcal{V})$ is a vector in $\mathbb{R}^{d_\text{visual}}$, output by a pretrained CNN truncated at the penultimate layer, $W_V$ is a projection matrix ($\in \mathbb{R}^{d \times d_\text{visual}}$) and $b_V$ a bias vector  --- in our experiments, $d_\text{visual}={2048}$.
The probability that the image region $\mathcal{V}$ corresponds to the class $i$ is set to be proportional to the cosine similarity between the projection $f_{\theta_V}(\mathcal{V})$ of $\mathcal{V}$ and the semantic representation $w_i$ of $i$:
\vspace{-0.1cm}
\begin{equation}
\log P( \mathcal{V}|i;\theta_V) \propto \cos(f_{\theta_V}(\mathcal{V}), w_i) := \log \widetilde{P}_\textit{visual}
\label{eq_visual} \end{equation}
\vspace{-0.7cm}

\paragraph{Context component}

The context component models $P(\mathcal{C}|i)$ by computing a compatibility score between the visual context $\mathcal{C}$, and the semantic representation $w_i$ of class~$i$.
More precisely, the conditional probability is written:
\vspace{-0.1cm}
\begin{align}
    \log P(\mathcal{C}|i;\theta_C) \propto f_{\theta_C}(\mathcal{C},w_i) &= f_{\theta^1_C} \big( h_{\theta^2_C}(\mathcal{C})\oplus w_i \big) \notag \\ 
    &:= \log \widetilde{P}_\textit{context}
    \label{eq_context}
\end{align}
\vspace{-0.7cm}

where $h_{\theta^2_C}(\mathcal{C}) \in \mathbb{R}^d$ is a vector representing the context, $\theta_C = \{\theta^1_\mathcal{C} ; \theta^2_\mathcal{C}\}$ are parameters to learn, and $\oplus$ is the concatenation operator.
To take non-linear and high-order interactions between $h_{\theta^2_C}(\mathcal{C})$ and $w_i$ into account, $f_{\theta_C^1}$ is modeled by a 2-layer Perceptron. 
We found that concatenating $h_{\theta_C^2}(\mathcal{C})$ with $w_i$ leads to better results than a cosine similarity, as done in equation~\ref{eq_visual} for the visual component.

To specify the modeling of $h_{\theta^2_C}(\mathcal{C})$, 
we propose various \textit{context models} depending on which context objects are considered and how they are represented.
Specifically, a context model is characterized by (a) the domain of context objects that are considered (i.e.\ source $\mathcal{S}$ or target $\mathcal{T}$) and (b) the way these objects are represented, either by a textual representation of their class label or by a visual representation of their image regions.
Accordingly, we distinguish: \\
$\bullet$ The \textit{low-level} ($L$) approach that computes a representation from the image region $\mathcal{V}_k$ of a context object. This produces the following context models: 
\begin{eqnarray*}
S_L= \{W_C \texttt{CNN}(\mathcal{V}_k)+b_C | k \in \mathcal{C} \cap \mathcal{S} \} \\
T_L= \{W_C \texttt{CNN}(\mathcal{V}_k)+b_C | k \in \mathcal{C} \cap \mathcal{T} \}
\end{eqnarray*}
$\bullet$ The \textit{high-level} ($H$) approach which considers semantic representations $w_k$ of the class labels $k$ of the context objects (only available for entities of the source domain). This produces context models: 
\begin{eqnarray*}
S_H= \{w_k | k \in \mathcal{C} \cap \mathcal{S}\} \mbox{ and } T_H= \{w_k | k \in \mathcal{C} \cap \mathcal{T} \}
\end{eqnarray*}
Note that $T_H$ is not defined in the zero-shot setting, since class labels of objects from the target domain are unknown; yet it is used to define Oracle models (section~\ref{oracles}).

%
These four basic sets of vectors can further be combined in various ways to form new context models (for instance: $S_L \cup T_L,S_H \cup S_L, S_H\cup S_L \cup T_L$, etc.). 
At last, $h_{\theta_C^2}$ averages the representations of these vectors to build a global context representation.
For example, $h_{\theta_C^2}(\mathcal{C}_{S_H \cup T_L})$ equals:
\vspace{-0.2cm}
\begin{equation*}
 \frac{ 1 }
{|\mathcal{C}_\mathcal{S}|  +  |\mathcal{C}_\mathcal{T}|}
\Big[
\sum\limits_{ (i, \mathcal{V}_i) \in \mathcal{C}_\mathcal{S}} 
\hspace{-0.35cm} w_i  + 
\sum\limits_{(j, \mathcal{V}_j) \in \mathcal{C}_\mathcal{T}}
\big( W_C.\texttt{CNN}(\mathcal{V}_j)+b_C \big)
\Big] 
\end{equation*}
\vspace{-0.2cm}
where $|\cdot|$ denotes the cardinality of a set of vectors.



\paragraph{Prior component}
The goal of the prior component is to assess whether an entity is frequent or not in images. We estimate $P(i)$ from the semantic representation $w_i$ of class~$i$:

\vspace{-0.5cm}
\begin{equation}
    \log P(i;\theta_P) \propto f_{\theta_P}(w_i) := \log \widetilde{P}_\textit{prior}
    \label{eq_prior}
\end{equation}
where $f_{\theta_P}$ is a 2-layer Perceptron that outputs a scalar. 

\subsection{Learning}
\label{learning}

In this section, we explain how we learn the energy functions $f_{\theta_C}$, $f_{\theta_V}$ and $f_{\theta_P}$. 
Each component (resp. context, visual, prior) of our model is assigned a training objective (resp. $\mathcal{L}_{C}$, $\mathcal{L}_{V}$, $\mathcal{L}_{P}$). As the components are independent by design, they are learned separately.
This allows for a better generalization in the target domain, as shown experimentally (section~\ref{scenarios}).
Besides, ensuring that some configurations are more likely than others motivates us to model each objective by a max-margin ranking loss, in which a positive configuration is assigned a lower energy than a negative one, following the \textit{learning to rank} paradigm \cite{wsabie}. 
Unlike previous works \cite{devise}, which are generally based on balanced datasets such as ImageNet and thus are not concerned with prior information, we want to avoid any bias coming from the imbalance of the dataset in $\mathcal{L}_C$ and $\mathcal{L}_V$, and learn the prior separately with $\mathcal{L}_P$. In other terms, the visual (resp. context) component should focus exclusively on the visual appearance (resp. visual context) of objects.
This is done with a careful sampling strategy of the negative examples within the ranking objectives, that we detail in the following.
To the best of our knowledge, such a discussion relative to prior modeling in learning objectives --- which is, in our view, paramount in imbalanced datasets such as Visual Genome --- has not been done in previous research. 

Positive examples are sampled among entities of the source domain from the data distribution $P^\star$: they consist in a single object for $\mathcal{L}_{P}$, an object/box pair for $\mathcal{L}_{V}$, an object/context pair for $\mathcal{L}_{C}$.  
To sample negative examples $j$ from the source domain, we distinguish two ways:

(1) For the prior objective $\mathcal{L}_{P}$, negative object classes are sampled from the \textit{uniform} distribution $U$: 
    
\vspace{-0.4cm}
\begin{equation}
 \mathcal{L}_{P} =
 \mathop{\mathbb{E}}\limits_{i \sim P^\star} \mathop{\mathbb{E}}\limits_{j \sim U}
 \big\lfloor \gamma_P - f_{\theta_P}(w_i)  + f_{\theta_P}(w_j) \big\rfloor_+
\end{equation}
\vspace{-0.5cm}



Noting $\Delta_{ji}:=f_{\theta_P}(w_j) - f_{\theta_P}(w_i)$, the contribution of two given objects $i$ and $j$ to this objective is:
\vspace{-0.05cm}
\begin{align*}
P^\star (i)\big\lfloor \gamma_P + \Delta_{ji} \big\rfloor_{+} 
+
P^\star (j)\big\lfloor \gamma_P - \Delta_{ji}  \big\rfloor_{+}
\end{align*} 
%
If $P^\star(i)>P^\star(j)$, i.e. when object class $i$ is more frequent than object class $j$, this term is minimized when  $\Delta_{ji}=-\gamma_P$, i.e. $f_{\theta_P}(w_i) = f_{\theta_P}(w_j) + \gamma_P > f_{\theta_P}(w_j)$. Thus, $\widetilde{P}_\textit{prior}(.;\theta_P)$ captures prior information, as it learns to rank objects based on their frequency.

(2) For the visual and context objectives, negative object classes are sampled from the prior distribution $P^\star(.)$: 


\vspace{-0.65cm}
\begin{align}
  \mathcal{L}_{V} &= \hspace{-0.14cm} \mathop{\mathbb{E}}\limits_{i, \mathcal{V} \sim P^\star} \hspace{-0.05cm} \mathop{\mathbb{E}}\limits_{j \sim P^\star}
   \big\lfloor \gamma_V \hspace{-0.05cm} - \hspace{-0.05cm} f_{\theta_V}(\mathcal{V})^\top w_i  + f_{\theta_V}(\mathcal{V})^\top w_j \big\rfloor_{\hspace{-0.04cm} +} \hspace{-0.3cm} \label{L_V}
   \\
  \mathcal{L}_{C} &= \hspace{-0.14cm} \mathop{\mathbb{E}}\limits_{i, \mathcal{C} \sim P^\star}  \mathop{\mathbb{E}}\limits_{j \sim P^\star} \big\lfloor \gamma_C \hspace{-0.05cm} - \hspace{-0.05cm} f_{\theta_C}\big(\mathcal{C}, w_i\big)  \hspace{-0.05cm}  +  \hspace{-0.05cm} f_{\theta_C}\big(\mathcal{C}, w_j\big)  \hspace{-0.05cm} \big\rfloor_{\hspace{-0.02cm} +}
\end{align}

\vspace{-0.3cm}
Similarly, the contribution of two given objects $i$, $j$ and a context $\mathcal{C}$ to the objective  $\mathcal{L}_{C}$ is: 
\vspace{-0.1cm}
\begin{align*} P^\star(i)P^\star(j)
\Big[P^\star(\mathcal{C}|i) \big\lfloor
 \gamma_V + f_{\theta_C}(\mathcal{C},w_j) - f_{\theta_C}(\mathcal{C},w_i) \big\rfloor_{+} \\
+ P^\star (\mathcal{C}|j)\big\lfloor \gamma_V  + f_{\theta_C}(\mathcal{C},w_i) - f_{\theta_C}(\mathcal{C},w_j)  \big\rfloor_+ \Big] \end{align*}
\vspace{-0.6cm}

Minimizing this term does not depend on the relative order between $P^\star(i)$ and $P^\star(j)$; thus, $\widetilde{P}_\textit{context}(\mathcal{C}|.;\theta_C)$ does not take prior information into account. Moreover, ${P^\star(\mathcal{C}|i)>P^\star(\mathcal{C}|j)}$ implies that $f_{\theta_C}(\mathcal{C},w_i)> f_{\theta_C}(\mathcal{C},w_j)$. 


The alternative, as done in DeViSE \cite{devise}, is to sample negative classes uniformly in the source domain in the objective $\mathcal{L}_V$. Thus, if the prior is uniform, DeViSE directly models $P(.|\mathcal{V})$; otherwise, $\mathcal{L}_V$ cannot be analyzed straightforwardly. Besides, the contributions of visual and prior information are mixed.
However, we show that learning the prior separately and imposing the context (resp. visual) component to exclusively focus on contextual (resp. visual) information is more efficient (section \ref{scenarios}).

\vspace{-0.2cm}
\subsection{Inference}
\label{inference}

\newcommand{\bigCI}{\mathrel{\text{\scalebox{1.07}{$\perp\mkern-10mu\perp$}}}}

In this section, we detail the inference process. 
The goal is to combine the predictions of the individual components of the model to form the global probability distribution $P(.|\mathcal{V},\mathcal{C})$.
In section~\ref{learning}, we detailed how to learn the functions $f_{\theta_C}$, $f_{\theta_V}$ and $f_{\theta_P}$, from which $ \log \widetilde{P}_\textit{context}$, $\log \widetilde{P}_\textit{visual}$ and $\log \widetilde{P}_\textit{prior}$ are deduced respectively.
However, the normalization constants in equations \ref{eq_visual}, \ref{eq_context}  and \ref{eq_prior}, which depend on the object class $i$ in the general case, are unknown.
As a simplifying hypothesis, we suppose that these normalization constants are scalars that we respectively note $\alpha_C$, $\alpha_V$ and $\alpha_P$. This leads to:

\vspace{-0.65cm}
\begin{equation}
P(.|\mathcal{V},\mathcal{C}) =
\underbrace{(\widetilde{P}_{\textit{context}})^{\alpha_C}}_{\textstyle P(\mathcal{C}|.)} . 
\underbrace{(\widetilde{P}_{\textit{visual}})^{\alpha_V}}_{\textstyle P(\mathcal{V}|.)} .
\underbrace{(\widetilde{P}_{\textit{prior}})^{\alpha_P}}_{\textstyle P(.)}
\end{equation}
\vspace{-0.6cm}

\begin{figure}[t]
 \centering
     \includegraphics[width=\linewidth]{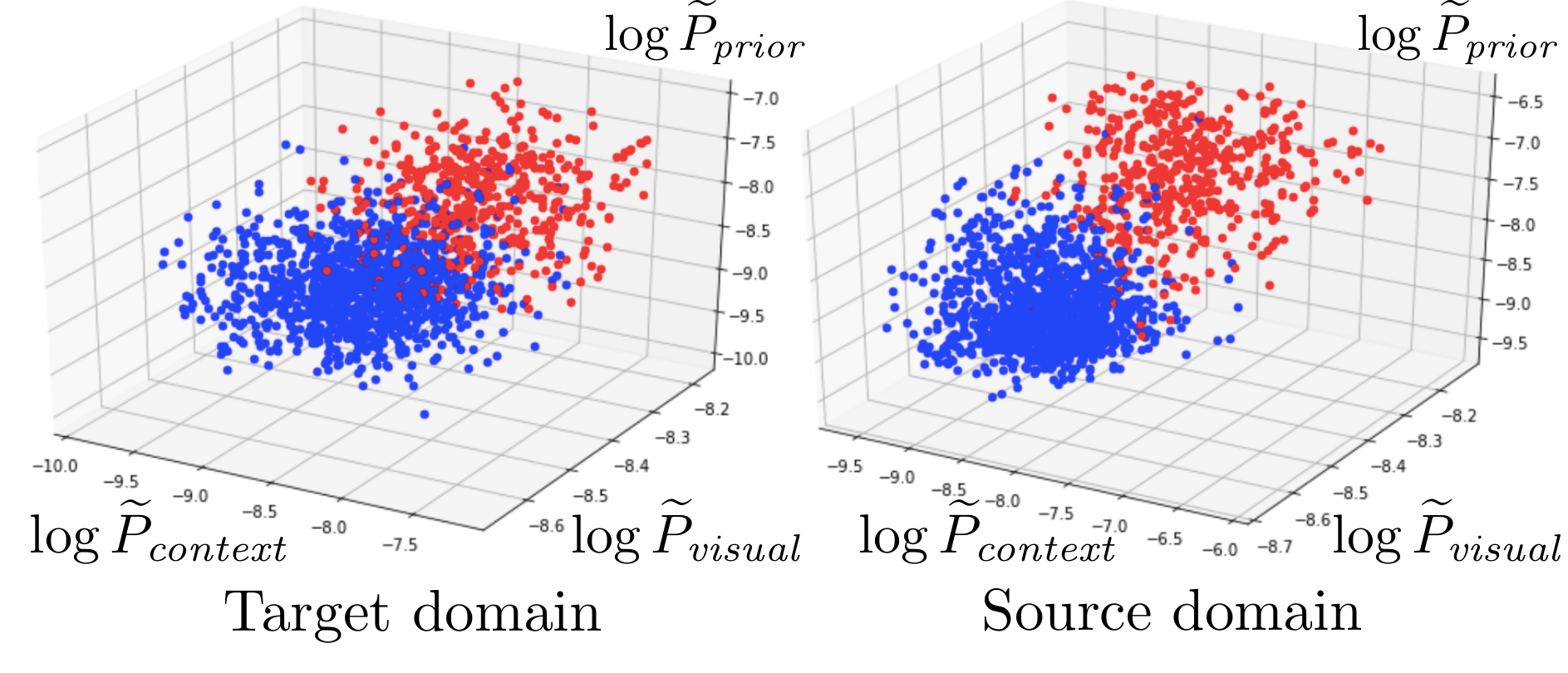}
     \vspace{-0.8cm}
     \caption{3D visualization of the unnormalized log-probabilities of each component ($N=500$). Context model ${S_L \cup S_H \cup T_L}$.}
     \vspace{-0.7cm}
     \label{3D}
 \end{figure}

To see whether this hypothesis is reasonable, we did some \textit{post-hoc} analysis of one of our model, and plotted
in Figure~\ref{3D} the values $\log \widetilde{P}_\textit{visual}$, $\log \widetilde{P}_\textit{context}$ and $\log \widetilde{P}_\textit{prior}$ for positive (red points) and negative (blue points) configurations $(i,\mathcal{V},\mathcal{C})$ of the test set of Visual Genome. We observe
that positive and negative triplets are well separated, which empirically validates our initial hypothesis.

Hyper-parameters $\alpha_C,\alpha_V$ and $\alpha_P$ are selected on the validation set to compute $P(.|\mathcal{C},\mathcal{V})$.
To build models that do not use a visual/contextual component, we simply 
select a subset of the probabilities and their respective hyperparameters. For example, $P(.|\mathcal{C}) = (\widetilde{P}_\textit{context})^{\alpha_C} (\widetilde{P}_\textit{prior})^{\alpha_P}$. 




\section{Experimental protocol}
\label{sec:counting:analysis}

\subsection{Data}

To measure the role of context in ZSL, a dataset that presents annotated objects within a rich visual context is required.
However, traditional ZSL datasets, such as AwA \cite{zsldatasetawa}, 
CUB-200 \cite{zsldatasetbird} or LAD \cite{zsldatasetlad}, are made of images that contain a unique object each, with no or very little surrounding visual context.
We rather use Visual Genome \cite{visualgenome}, a large-scale image dataset (108K images) annotated at a fine-grained level (3.8M object instances), covering various concepts (105K unique object names). This dataset is of particular interest for our work, as objects have richly annotated contexts ($31$ object instances per image on average). 
In order to shape the data to our task, we randomly split the set of images of Visual Genome into train/validation/test sets (70\%/10\%/20\% of the total size).
To build the set $\mathcal{O}$ of all objects classes, we select classes which appear at least 10 times in Visual Genome and have an available \textit{Word2vec} representation. 
$\mathcal{O}$ contains $4842$ object classes; it amounts to 3.4M object instances in the dataset. This dataset is highly imbalanced as 10\% of most represented classes amount to 84\% of object instances.
We define the \textit{level of supervision} $p_\text{sup}$ as the ratio of the size of the source domain over the total number of objects: $p_\text{sup} = |\mathcal{S}|/|\mathcal{O}|$.
For a given $p_\text{sup}$ ratio, the source $\mathcal{S}$ and target $\mathcal{T}$ domains are built by randomly splitting $\mathcal{O}$ accordingly. 
Every object is annotated with a bounding box and we use this supervision in our model for entities of both source and target domains. 
To facilitate future work on context-aware ZSL, we publicly release data splits and annotations \footnote{\url{https://data.lip6.fr/context_aware_zsl/}}.



\vspace{-0.2cm}
\subsection{Evaluation methodology and metrics}
\label{mfr}

We adopt the conventional setting for ZSL, which implies entities to be retrieved only among the target domain $\mathcal{T}$. Besides, we also evaluate the performance of the model to retrieve entities of the source domain $\mathcal{S}$ (with models tuned on the target domain).

The model's prediction takes the form of a list of $n$ classes, sorted by probability;
the rank of the correct class in that list is noted $r$. Depending on the setting, $n$ equals $|\mathcal{T}|$ or $|\mathcal{S}|$.
We define the First Relevant (FR) metric with 
${\text{FR} = \frac{2}{n-1}(r-1)}$.
To further evaluate the performance over the whole test set, the Mean First Relevant (MFR) metric is used \cite{DBLP:journals/sigir/Fuhr17}. It is computed by taking the mean value of FR scores obtained on each image of the test set.
Note that the factor $\frac{2}{n-1}$ rescales the metric such that the MFR score of a random baseline is 100\%, while the MFR of a perfect model would be 0\%.
The MFR metric has the advantage to be interval-scale-based, unlike more traditional Recall@$k$ metrics or Mean Reciprocal Ranks metrics \cite{DBLP:conf/ictir/FerranteFP17}, and thus can be averaged; this allows for meaningful comparison with a varying $p_\text{sup}$.



\vspace{-0.1cm}
\subsection{Scenarios and Baselines}
\label{oracles}
\paragraph{Model scenarios}
Model scenarios depend on the information that is used in the probabilistic setting: $\varnothing$, $\mathcal{C}$, $\mathcal{V}$ or both $\mathcal{C}$ and $\mathcal{V}$.
When contextual information is involved, a context model $\star$ is specified to represent $\mathcal{C}$, which we note $\mathcal{C}_{\star}$. The different context models are $\star \in \{S_H, S_L, T_L, S_L \cup T_L, S_H \cup T_L, S_L \cup S_H \cup T_L\}$.
For clarity's sake, we note our model M.
For example, M$(\mathcal{C}_{S_H \cup T_L},\mathcal{V})$ models the probability $P(\mathcal{C}_{S_H \cup T_L}|.)P(\mathcal{V}|.)P(.)$ as explained in \ref{inference}, M$(\mathcal{V})$ models $P(\mathcal{V}|.)P(.)$, and M$(\varnothing)$ models $P(.)$.

\vspace{-0.3cm}
\paragraph{Oracles} To evaluate upper-limit performances for our models, we define Oracle baselines where classes of target objects are used, which is not allowed in the zero-shot setting.
Note that every Oracle leverages visual information. \\
$\bullet$ \emph{True Prior:} This Oracle uses, for its prior component, the true prior distribution $P^\star(i)=\frac{\#i}{M}$ computed for all objects of both source and target domains on the full dataset, where $\#i$ is the number of instances of the $i$-th class in images and $M$ is the total number of images. 
\\
$\bullet$ \emph{Visual Bayes:} This Oracle uses $P^\star(.)$ for its prior component as well. Its context component uses co-occurrence statistics between objects computed on the full dataset:
$P^\text{im}(\mathcal{C}|i)=\prod_{c \in \mathcal{C}}P_\text{co-oc}(c|i)$
where $P_\text{co-oc}(c|i)=\frac{\#(c,i)M}{\#c\#i}$ is the probability that objects $c$ and $i$ co-occur in images, with $\#(c,i)$ the number of co-occurrences of $c$ and $i$. \\
$\bullet$ \emph{Textual Bayes:} Inspired by \cite{DBLP:journals/corr/BengioDEILRSS13}, this Oracle is similar to Visual Bayes, except that its prior $P^\text{text}(.)$ and context component $P^\text{text}(.|\mathcal{C})$ are based on textual co-occurrences instead of image co-occurrences:  $P_\text{co-oc}(c|i)$ is computed by counting co-occurrences of words $c$ and $i$ in windows of size 8 in the Wikipedia dataset, and $P^\text{text}(i)$ is computed by summing the number of instances of the $i$-th class divided by the total size of Wikipedia. \\
$\bullet$ \emph{Semantic representations for all objects:} M$(\mathcal{C}_{S_H \cup T_H},\mathcal{V})$ uses word embeddings of both source and target objects. 

\textbf{Baselines} 
\\
$\bullet$ M$(\mathcal{C}\oplus \mathcal{V}$): To study the validity of the hypothesis about the conditional independence of $\mathcal{C}$ and $\mathcal{V}$, we introduce a baseline where we directly model $P(\mathcal{C},\mathcal{V}|.)P(.)$. To do so, we replace, in the expression of $\mathcal{L}_V$ (equation \ref{L_V}), $f_{\theta_V}(\mathcal{V})$ by the concatenation of $h(\mathcal{C})$ and $f_{\theta_V}(\mathcal{V})$ projected in $\mathbb{R}^d$ with a 2-layer Perceptron.
\\
$\bullet$ DeViSE$(\mathcal{V})$: \hspace{-4pt} To evaluate the impact of our Bayesian model (equation \ref{p_global}) and our sampling strategy (section \ref{learning}), we compare against DeViSE \cite{devise}. DeViSE$(\mathcal{V})$ is different from M$(\mathcal{V})$ because negative examples in $\mathcal{L}_V$ are uniformly sampled, and the prior $P(.)$ is not learned.
\\
$\bullet$ DeViSE$(\mathcal{C}\oplus \mathcal{V})$: similarly to M$(\mathcal{C}\oplus \mathcal{V})$, we define a baseline that does not rely on the conditional independence of $\mathcal{C}$ and $\mathcal{V}$, using the same sampling strategy as DeViSE.
\\
$\bullet$ M$(\mathcal{C}_I, \mathcal{V})$:
To understand the importance of context supervision, i.e. annotations of context objects (boxes and classes), we design a baseline where no context annotations are used. The context is the whole image without the zone $\mathcal{V}$ of the object, which is masked out.
The associated context model is $\star=I$ with $ h(\mathcal{C_I}) = g_{\theta_I} (I \setminus \mathcal{V}) $ ; $g_{\theta_I}$ is a parametric function to be learned. This baseline is inspired from \cite{DBLP:journals/cacm/TorralbaMF10}, where global image features are used to refine the prediction of an image model. 

\vspace{-0.2cm}
\subsection{Implementation details}

For each objective $\mathcal{L}_{C}, \mathcal{L}_{V}$ and $\mathcal{L}_{P}$, 
at each iteration of the learning algorithm, 5 negative entities are sampled per positive example.
Word representations are vectors of $\mathbb{R}^{300}$, learned with the Skip-Gram algorithm \cite{mikolov2013} on Wikipedia.
Image regions are cropped, rescaled to (299$\times$299), and fed to \texttt{CNN}, an Inception-v3 CNN \cite{inception}, whose weights are kept fixed during training. This model is pretrained on ImageNet \cite{imagenet}. As a result, every ImageNet class that belongs to the total set of objects $\mathcal{O}$ was included in the source domain $\mathcal{S}$. 
Models are trained with Adam \cite{adam} and regularized with a L2-penalty; the weight of this penalty decreases when the level of supervision increases, as the model is less prone to overfitting.
All hyper-parameters are cross-validated on classes of the target domain, on the validation set. 

\section{Results}
\label{results}

\begin{table}[t]
    \centering
        \caption{Evaluation of various information sources, with varying levels of supervision. MFR scores in $\%$. $\delta_C$ is the relative improvement (in $\%$) of M$(\mathcal{C}_{S_H}, \mathcal{V})$ over M$(\mathcal{V})$.}
        \vspace{5.5pt}
    \begin{tabularx}{\columnwidth}{@{\hspace{0.1pt}} p{0.01pt} @{\hspace{4pt}}  >{\raggedleft}p{45pt} @{\hspace{2.5pt}} | M M M | M M M} 
        & & \multicolumn{3}{c}{Target domain $\mathcal{T}$} & \multicolumn{3}{c}{Source domain $\mathcal{S}$} \\
        & $p_\text{sup}$ & \multicolumn{1}{c}{10\%} & \multicolumn{1}{c}{50\%} & \multicolumn{1}{c|}{90\%} & \multicolumn{1}{c}{10\%} & \multicolumn{1}{c}{50\%} & \multicolumn{1}{c}{90\%} \\
        
        & {\hspace{-10pt} \fontsize{8}{0}\selectfont Domain size} & \multicolumn{1}{c}{\fontsize{9}{0}\selectfont 4358} & \multicolumn{1}{c}{\fontsize{9}{0}\selectfont 2421} & \multicolumn{1}{c|}{\fontsize{9}{0}\selectfont 484} & \multicolumn{1}{c}{\fontsize{9}{0}\selectfont 484} & \multicolumn{1}{c}{\fontsize{9}{0}\selectfont 2421} & \multicolumn{1}{c}{\fontsize{9}{0}\selectfont 4358} \\
        
        \hline
        \multirow{5}{*}{\rotatebox[origin=c]{90}{\fontsize{9pt}{0}\selectfont \textbf{Models}}} & \textit{Random} & \textit{100} & \textit{100} & \textit{100} & \textit{100} & \textit{100} & \textit{100} \\
        & M$(\varnothing)$  & 38.6 & 23.7 & 13.8& 12.0 & 10.6 & 11.2 \\ 
        & M$(\mathcal{V})$ & 20.5 & 10.7 & 6.0& 1.5 & 2.6 & 3.6\\
        & M$(\mathcal{C}_{S_H})$ & 28.7 & 14.4 & 9.1& 4.2 & 4.3 & 4.4  \\
        & M$(\mathcal{C}_{S_H}, \mathcal{V})$ & \textbf{18.1} & \textbf{9.0} & \textbf{5.2} & \textbf{1.1} & \textbf{1.9} & \textbf{2.4} \\
        \hline
        & $\delta_\mathcal{C}$ (\%) & \textit{11.6} & \textit{16.4} & \textit{12.1}& \textit{23.7} & \textit{27.3} & \textit{31.5} 
            
    \end{tabularx}
    
    \label{tab:various1}
    \vspace{-14pt}
\end{table}

\subsection{The importance of context}
\label{importance_context}

In this section, we evaluate the contribution of contextual information, with varying levels of supervision $p_\text{sup}$.
We fix a simple context model ($\star=S_H$) and report MFR results with $p_\text{sup}=10,50,90\%$ in Table~\ref{tab:various1} for every combination of information sources: $\varnothing$, $\mathcal{V}$, $\mathcal{C}$ and $(\mathcal{C},\mathcal{V})$ --- we observe similar trends for the other context models.
Results highlight that contextual knowledge acquired from the source domain can be transferred to the target domain, as M$(\mathcal{C}_{S_H})$ significantly outperforms the \textit{Random} baseline.
As expected, it is not as useful as visual information: M$(\mathcal{V}) \s $ M$(\mathcal{C}_{S_H})$, where $\s$ means lower MFR scores, i.e.\ better performances.
However, Table~\ref{tab:various1} demonstrates that contextual and visual information are complementary: using M$(\mathcal{C}_{S_H},\mathcal{V})$ outperforms both M$(\mathcal{C}_{S_H})$ and M$(\mathcal{V})$.
Interestingly, as the learned prior model M$(\varnothing)$ is also able to generalize, we show that visual frequency can somehow be learned from textual semantics, which extends previous work where word embeddings were shown to be a good predictor of textual frequency \cite{frequence}. 

When $p_\text{sup}$ increases, we observe that all models are better at retrieving objects of the target domain (i.e. MFR decreases), which is intuitive because models are trained on more data and thus generalize better to recognize entities from the target domain.
Besides, when $p_\text{sup}$ increases, the context is also more abundant. This explains: (1) the decreasing MFR values for model M$(\mathcal{C}_{S_H})$ on $\mathcal{T}$, (2) the increasing relative improvement $\delta_C$ of M$(\mathcal{C}_{S_H},\mathcal{V})$ over M$(\mathcal{V})$ on $\mathcal{S}$.
However, on the target domain, we note that $\delta_C$ does not monotonously increase with $p_\text{sup}$. 
A possible explanation is that the visual component improves faster than the context component, so the relative contribution brought by context to the final model M$(\mathcal{C}_{S_H}, \mathcal{V})$ decreases after $p_\text{sup}=50\%$. 
Since the highest relative improvement $\delta_C$ (in $\mathcal{T}$) is attained with $p_\text{sup}=50\%$, we fix the standard level of supervision $p_\text{sup}=50\%$ in the rest of the experiments; this amounts to 2421 classes in both source and target domains.

\vspace{-0.2cm}
\subsection{Modeling contextual information}
\label{scenarios}


\newlength{\lfontnb}
\setlength{\lfontnb}{8pt}
\newcommand*\rota{\multicolumn{1}{R{0}{0pt}}}
\begin{table}[t]
    \centering
    \small
        \caption{MFR performances (given in $\%$) for all baselines and scenarios. $p_\text{sup}=50\%$.
       Oracle results, written in italics, are not taken into account to determine the best scores, written in bold. 
       }
        \begin{tabularx}{\columnwidth}{@{\hspace{0pt}} c @{\hspace{2.5pt}} l @{\hspace{5pt}} r @{\hspace{0.5pt}} | @{\hspace{2pt}} r @{\hspace{7pt}} @{\hspace{2pt}} r }
    & {Model} & Probability & \multicolumn{1}{c}{$\mathcal{T}$} & \multicolumn{1}{c}{$\mathcal{S}$} \\
        \hline
     \multirow{4}{*}{\rotatebox[origin=c]{90}{\textit{Oracles}}} & \textit{Textual Bayes} & {\fontsize{7.5}{0}\selectfont $P^\text{text}(\mathcal{C}|.)P(\mathcal{V}|.)P^\text{text}(.)$} & \hspace{2pt} \textit{14.54} & \textit{6.73} \\
    & {\fontsize{\lfontnb}{0}\selectfont M$(\mathcal{C}_{S_H \cup T_H}, \mathcal{V})$} & {\fontsize{7.5}{0}\selectfont $P(\mathcal{C}_{S_H \cup T_H}|.)P(\mathcal{V}|.)P(.)$} &  \textit{7.57}&  \textit{2.53}  \\
    & \textit{True Prior} & {\fontsize{7.5}{0}\selectfont $P(\mathcal{V}|.) P^\star(.)$} & \textit{4.92}&  \textit{2.63} \\
    & \textit{Visual Bayes} & {\fontsize{7.5}{0}\selectfont $P^\text{im}(\mathcal{C}|.)P(\mathcal{V}|.)P^\star(.)$} &   \textit{3.40} & \textit{2.11} \\
    
        \hline
    \multirow{4}{*}{\rotatebox[origin=c]{90}{\textbf{Baselines}}} & {\fontsize{\lfontnb}{0}\selectfont DeViSE$(\mathcal{V})$} & {\fontsize{7.5}{0}\selectfont $P(.|\mathcal{V})$} & 10.73 & 3.62 \\
    & {\fontsize{\lfontnb}{0}\selectfont DeViSE$(\mathcal{C}_{S_H} \oplus \mathcal{V})$ } & {\fontsize{7.5}{0}\selectfont$P(.|\mathcal{C}_{S_H}, \mathcal{V})$} &  10.11&  3.11 \\ 
    & {\fontsize{\lfontnb}{0}\selectfont M$(\mathcal{C}_{S_H} \oplus \mathcal{V})$} & {\fontsize{7.5}{0}\selectfont$P(\mathcal{C}_{S_H}, \mathcal{V}|.)P(.)$}  & 10.07 &  1.85 \\ 
    & {\fontsize{\lfontnb}{0}\selectfont M$(\mathcal{C}_{I}, \mathcal{V})$} & {\fontsize{7.5}{0}\selectfont $P(\mathcal{C}_{I}|.)P(\mathcal{V}|.)P(.)$} & 9.19 &  2.13 \\ 
    
        \hline
    \multirow{7}{*}{\rotatebox[origin=c]{90}{\textbf{Our models}}} & {\fontsize{\lfontnb}{0}\selectfont M$(\mathcal{V})$} & {\fontsize{7.5}{0}\selectfont $P(\mathcal{V}|.)P(.)$} & 10.72  & 2.64\\
     & {\fontsize{\lfontnb}{0}\selectfont M$(\mathcal{C}_{S_L}, \mathcal{V})$} & {\fontsize{7.5}{0}\selectfont $P(\mathcal{C}_{S_L}|.)P(\mathcal{V}|.)P(.)$} & 9.01 & 2.05 \\
    & {\fontsize{\lfontnb}{0}\selectfont M$(\mathcal{C}_{T_L}, \mathcal{V})$} & {\fontsize{7.5}{0}\selectfont $P(\mathcal{C}_{T_L}|.)P(\mathcal{V}|.)P(.)$}& 9.00 & 2.13 \\
    
    & {\fontsize{\lfontnb}{0}\selectfont M$(\mathcal{C}_{S_H}, \mathcal{V})$} & {\fontsize{7.5}{0}\selectfont $P(\mathcal{C}_{S_H}|.)P(\mathcal{V}|.)P(.)$} & 8.96  & 1.92\\ 
    
    & {\fontsize{\lfontnb}{0}\selectfont M$(\mathcal{C}_{S_H \cup S_L}, \mathcal{V})$} & {\fontsize{7.5}{0}\selectfont $P(\mathcal{C}_{S_H \cup S_L}|.)P(\mathcal{V}|.)P(.)$} & 8.71  & 1.88\\ 
    
    & {\fontsize{\lfontnb}{0}\selectfont M$(\mathcal{C}_{S_L \cup T_L}, \mathcal{V})$} &
    {\fontsize{7.5}{0}\selectfont $P(\mathcal{C}_{S_L \cup T_L}|.)P(\mathcal{V}|.)P(.)$} & 8.60& 1.93 \\ 
    
    & {\fontsize{\lfontnb}{0}\selectfont M$(\mathcal{C}_{S_H \cup T_L}, \mathcal{V})$} &
    {\fontsize{7.5}{0}\selectfont $P(\mathcal{C}_{S_H \cup T_L}|.)P(\mathcal{V}|.)P(.)$} & 8.52 & 1.86 \\ 
    
    & {\fontsize{\lfontnb}{0}\selectfont M$(\mathcal{C}_{S_H \cup S_L \cup T_L}, \mathcal{V})$} & {\fontsize{7.5}{0}\selectfont $P(\mathcal{C}_{S_H \cup S_L \cup T_L}|.)P(\mathcal{V}|.)P(.)$} & \textbf{8.31}&  \textbf{1.79}  \\ 
    
    \end{tabularx}
    \label{full_table}
    \vspace{-0.5cm}
\end{table}

In this section, we compare the different context models; results are reported in Table \ref{full_table}.
First, underlying hypotheses of our model are experimentally tested. (1) Modeling context and prior information with semantic representations (models M$(\mathcal{C}_\star, \mathcal{V}))$ is far more efficient than using direct textual co-occurrences, as shown by the \textit{Textual Bayes} baseline, which is the weaker model despite being an Oracle.
(2) Moreover, we show that the hypothesis on the conditional independence of $\mathcal{C}$ and $\mathcal{V}$ is acceptable, as separately modeling $\mathcal{C}$ and $\mathcal{V}$ gives better results than jointly modeling them (i.e.\ M$(\mathcal{C}_{S_H},\mathcal{V})$ $\s$ M$(\mathcal{C}_{S_H} \oplus \mathcal{V})$). (3) Furthermore, we observe that our approach M$(\mathcal{V}$) is more efficient to capture the imbalanced class distribution of the source domain, compared to DeViSE$(\mathcal{V})$; indeed, \textit{True Prior} $\approx$ M$(\mathcal{V})$ , whereas \textit{True Prior} $\s$ DeViSE$(\mathcal{V})$ on $\mathcal{S}$. Even if the improvement is only significant for the source domain $\mathcal{S}$, it indicates that separately using information sources is clearly a superior approach to further integrate contextual information. 

Second, as observed in the case of the context model $S_H$ (section~\ref{importance_context}), using contextual information is always beneficial. Indeed, all models with context M$(\mathcal{C}_\star, \mathcal{V})$ improve over M$(\mathcal{V})$ --- which is the model with no contextual information --- both on target and source domains.
In more details, we observe that performances increase when additional information is used: 
(1) when the bounding boxes annotations are available: all of our models that use both $\mathcal{C}$ and $\mathcal{V}$ outperform the baseline M$(\mathcal{C}_I,\mathcal{V})$, which could also be explained by the useless noise outside the object boxes in the image and the difficulty of computing a global context from raw image,
(2) when context objects are labeled and high-level features are used instead of low-level features, e.g.\ $S_H \s S_L$ and $S_H \cup T_H \s S_H \cup T_L$,
(3) when more context objects are considered (e.g.\ $S_L \cup T_L \s S_L$),
(4) when low-level information is used complementarily to high-level information (e.g.\ $S_L \cup S_H \cup T_L \s S_L \cup T_L$).
As a result, the best performance is attained for M$(\mathcal{C}_{S_L \cup S_H \cup T_L}, \mathcal{V})$, with a $22\%$ (resp. $32\%$) relative improvement in the target (resp. source) domain compared to M$(\mathcal{V})$.

We note that there is still room for improvement to approach ground-truth distributions for objects of the target domain (e.g, towards word embeddings able to better capture visual context). Indeed, even if our models outperform \textit{True Prior} and \textit{Visual Bayes} on the source domain, these Oracle baselines are still better on the target domain, hence showing that learning the visual context of objects from textual data is challenging. 

\vspace{-0.2cm}
\subsection{Qualitative Experiments}
\label{qualitative}

\begin{figure}[t]
\centering
    \includegraphics[width=\linewidth]{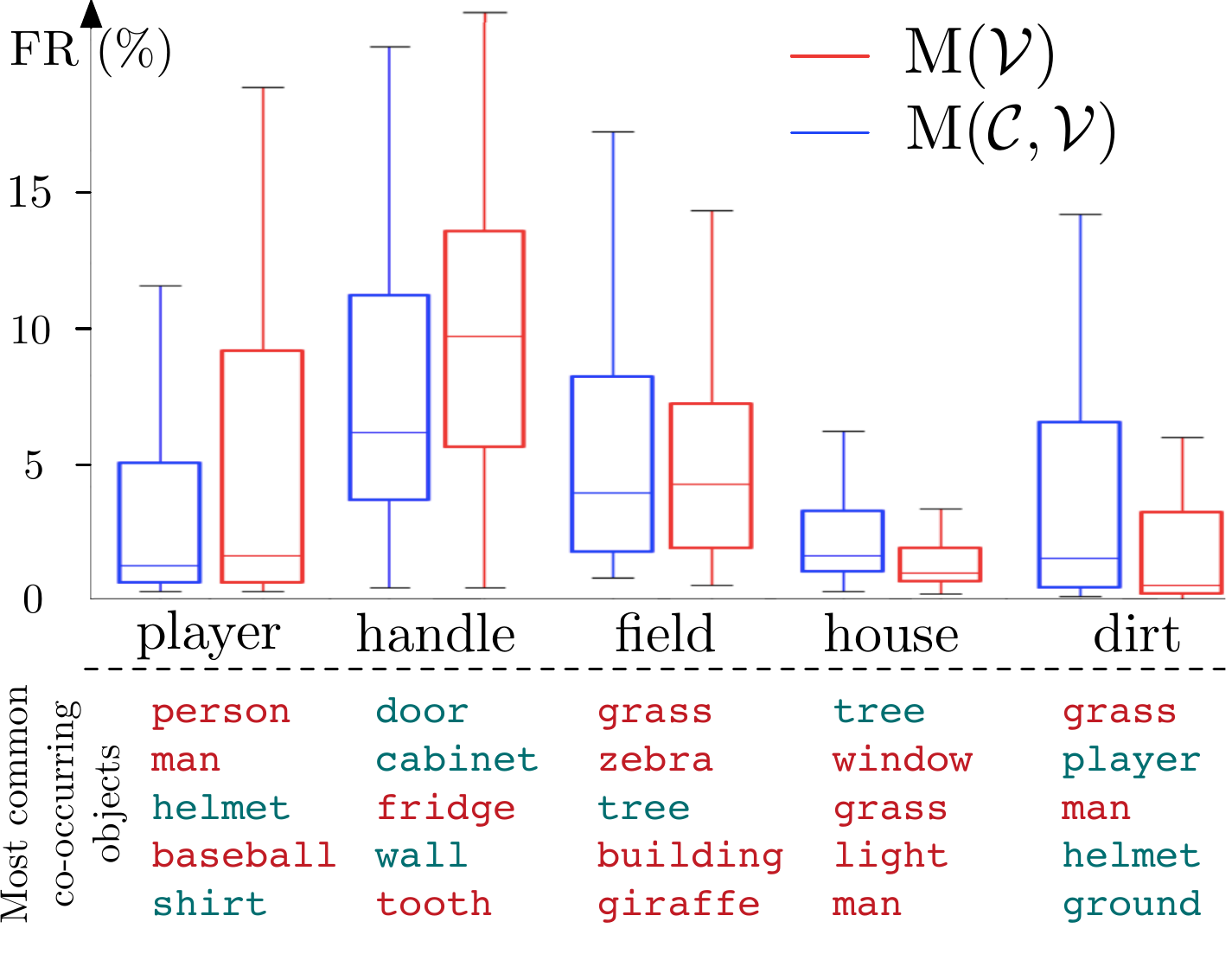}
   \vspace{-0.9cm}
    \caption{
    Boxplot representing the distribution of the correct ranks (First Relevant in $\%$) for five randomly selected classes of the target domain, with the context model $ S_L \cup  S_H \cup T_L$.
    Below are listed, by order of frequency, the classes that co-occur the most with the object of interest (classes of $\mathcal{T}$ in green; $\mathcal{S}$ in red). }
    \label{boxplot}
    \vspace{-0.6cm}
\end{figure}

To gain a deeper understanding of contextual information, we compare in Figure~\ref{boxplot} the predictions of M$(\mathcal{V})$ and the global model M$(\mathcal{C},\mathcal{V})$.
We randomly select five classes of the target domain and plot, for all instances of these classes in the test set of Visual Genome, the distribution of the predicted ranks of the correct class (in percentage); we also list the classes that appear the most in the context of these classes.
We observe that, for certain classes (\textit{player}, \textit{handle} and \textit{field}), contextual information helps to refine the predictions; for others (\textit{house} and \textit{dirt}), contextual information degrades the quality of the predictions.

First, we can outline that visual context can guide the model towards a more precise prediction. 
For example, a \textit{player}, without context, could be categorized as \textit{person}, \textit{man} or \textit{woman}; but visual context provides important complementary information (e.g, \textit{helmet}, \textit{baseball}) that grounds \textit{person} in a sport setting, and thus suggests that the \textit{person} could be playing.
Visual context is also particularly relevant when the object of interest has a generic shape. For example, \textit{handle}, without context, is visually similar to many round objects; but the presence of objects like \textit{door} or \textit{fridge} in the context helps determine the nature of the object of interest.

\begin{figure}[t]
\centering
    \includegraphics[trim={0 5.7cm 0 0},clip,width=\linewidth]{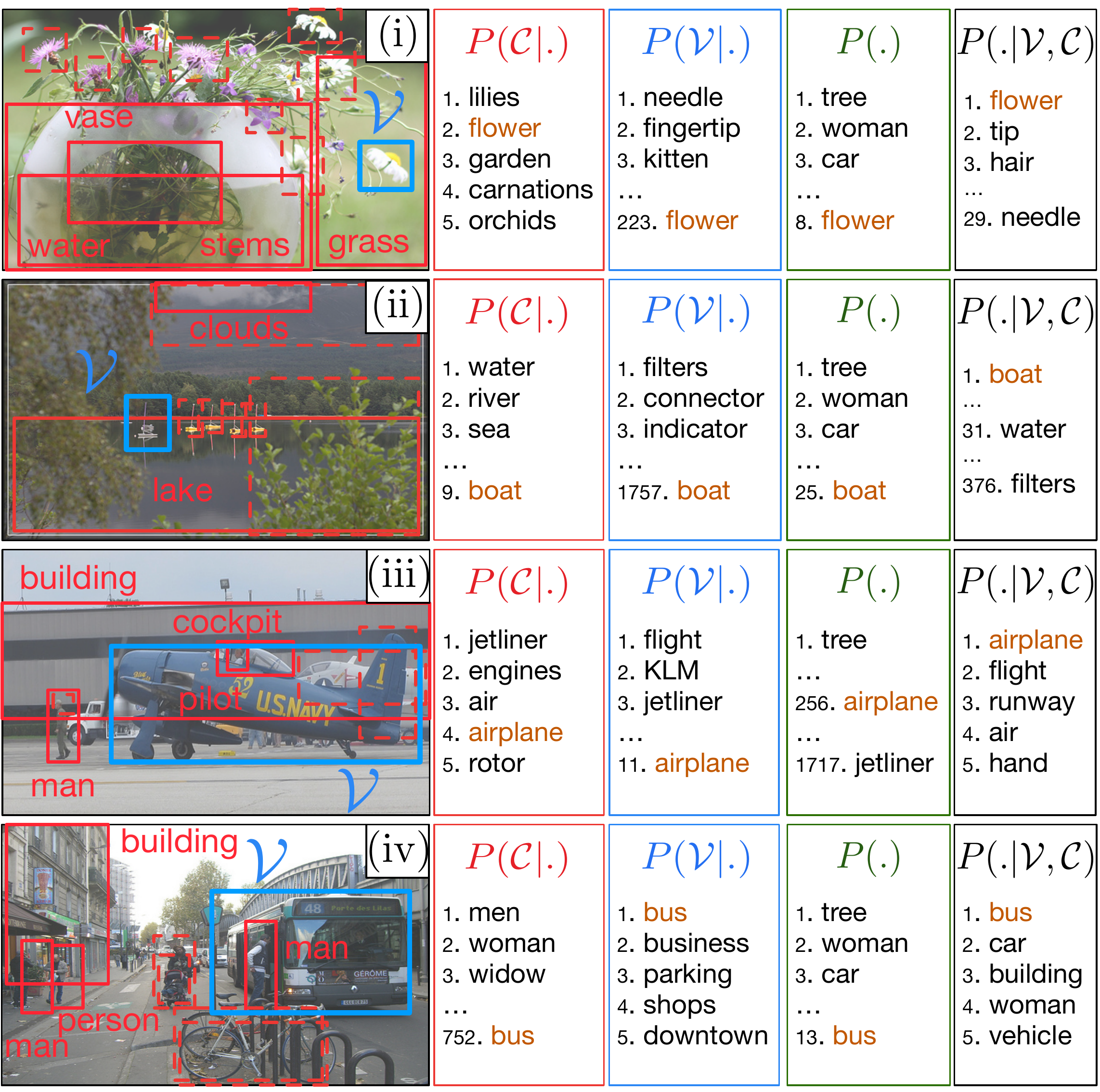}
    \vspace{-0.8cm}
    \caption{Qualitative examples where the global model M$(\mathcal{C}_{S_L \cup S_H \cup T_L},\mathcal{V})$ correctly retrieves the class ($\mathcal{T}$ classes only).}
    \label{quali}
    \vspace{-0.6cm}
\end{figure}

To get a better insight on the role of context, we cherry-picked examples where the visual or the prior component is inacurrate and the context component is able to counterbalance the final prediction (Figure~\ref{quali}). 
In (i), for example, the visual component ranks \textit{flower} at position 223. However, the context component assesses \textit{flower} to be highly probable in this context, due to the presence of source objects like \textit{vase}, \textit{water}, \textit{stems} or \textit{grass}, but also target objects like the other flowers around. 
At the inference phase, probabilities are aggregated and \textit{flower} is ranked first. 

It is worth noting that our work is not without limitations. Indeed, some classes (such as \textit{house} and \textit{dirt}) have a wide range of possible contexts; in these cases, context is not a discriminating factor. 
This is confirmed by a complementary analysis: the Spearman correlation between the number of unique context objects and $\delta_C$, the relative gain of M$(\mathcal{C}_{S_H},\mathcal{V})$ over M$(\mathcal{V})$, is $\rho=-0.31$. 
In other terms, contextual information is useful for specific objects, which appear in particular contexts; for objects that are too generic, adding contextual information can be a source of noise.

\vspace{-0.3cm}
\section{Conclusion}
In this paper, we introduced a new approach for ZSL: \textit{context-aware ZSL}, along with a corresponding model, using complementary contextual information that significantly improves predictions. 
Possible extensions could include spatial features of objects, and, more importantly, removing the dependence on the detection of object boxes to make it fully applicable to real-world images (e.g.\ by using a Region Proposal Network \cite{faster})
Finally, designing grounded word embeddings that include more visual context information would also benefit such models.


\section*{Acknowledgments}
This  work  is  partially  supported  by  the  CHIST-ERA  EU project MUSTER 1 (ANR-15-CHR2-0005) and the Labex SMART  (ANR-11-LABX-65)  supported  by  French  state funds  managed  by  the  ANR  within  the  Investissements d'Avenir program under reference ANR-11-IDEX-0004-02.

{
\bibliographystyle{icml2019} 
\bibliography{biblio}
}

%
\newpage
\appendix
\section{Additional negative results}
\begin{figure}[h]
    \centering
    \includegraphics[width=\linewidth]{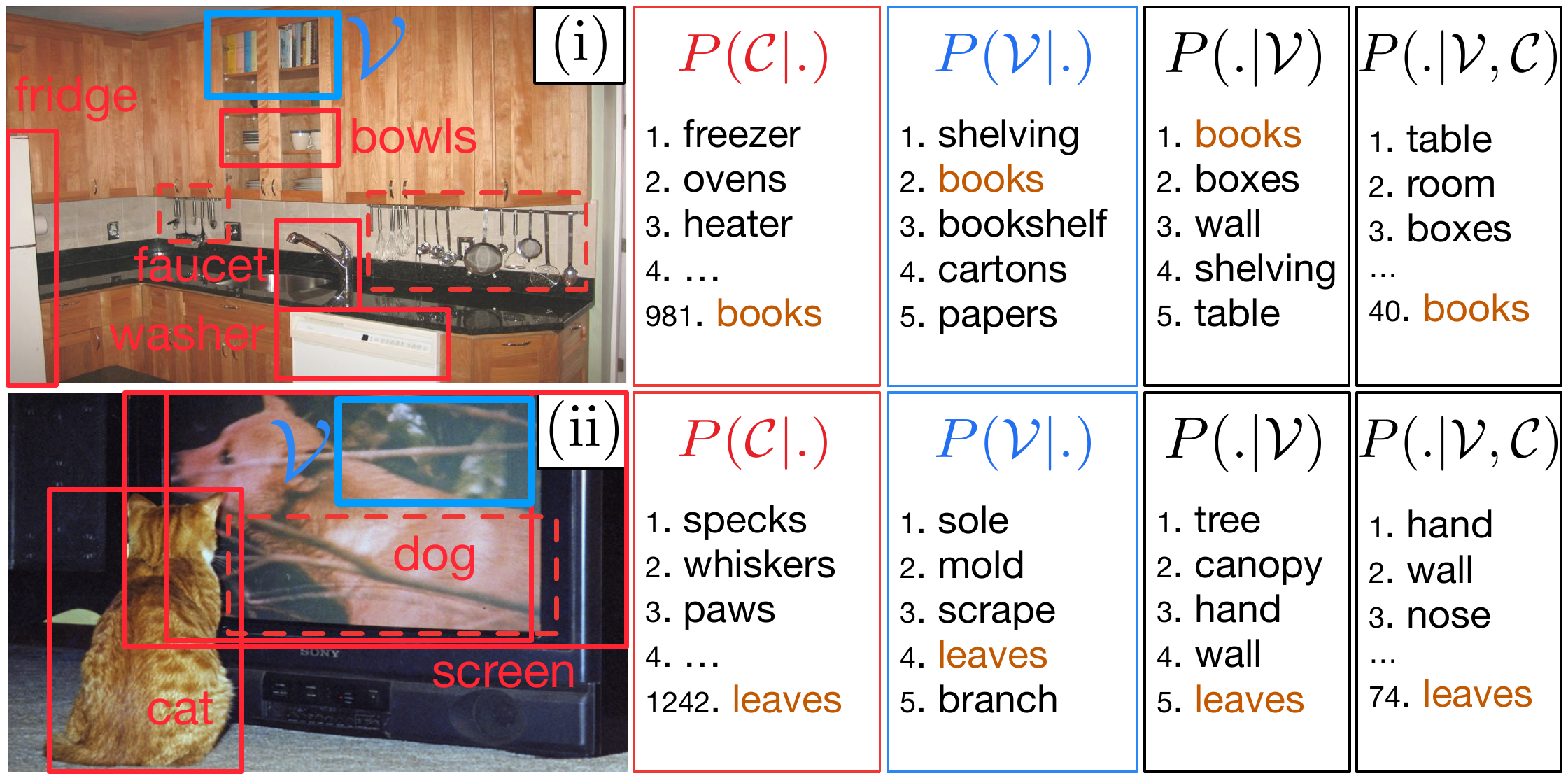}
    \vspace{-0.5cm}
    \caption{Qualitative analysis: negative examples where the use of the context leads to degraded predictions, i.e.\ examples where model M$(\mathcal{C}_{S_L \cup S_H \cup T_L},\mathcal{V})$ is worse than the simpler model M$(\mathcal{V})$ ($\mathcal{T}$ classes only).}
    \label{qualineg}
\end{figure}

As explained in Section~\ref{qualitative}, using contextual information can sometimes degrade predictions.
We provide here additional examples, when an object occurs in an environment in which it is unexpected.
For example, Figure~\ref{qualineg} shows a picture of a kitchen where the object of interest to be predicted is ``books''.
Given only the surrounding environment, predicted objects are logically related to the environment of a kitchen (``freezer'', ``oven'', \dots), and the correct label is badly ranked (because it is unexpected in such an environment).
However, the model M$(\mathcal{V})$ retrieves the correct label, given only the region of interest.
Finally, integrating contextual information in the final model M$(\mathcal{C}_{S_L \cup S_H \cup T_L}, \mathcal{V})$ leads to worse performances over M$(\mathcal{V})$.

\section{Generalized ZSL}
In the previous sections, retrieval in done only among classes of the domain of interest, this is the classical zero-shot learning setting. We now report results obtained when both source and target object classes exist in the retrieval space: this setting amounts to \textit{generalized zero-shot learning}. Results are reported in Table~\ref{generalized}.
\begin{table}[h]
    \centering
        \caption{\label{generalized}Evaluation of various information sources, with varying levels of supervision. Generalized ZSL setting. MFR scores in $\%$. $\delta_C$ is the relative improvement (in $\%$) of M$(\mathcal{C}_{S_H}, \mathcal{V})$ over M$(\mathcal{V})$.}
        \vspace{0.2cm}
    \begin{tabularx}{\columnwidth}{@{\hspace{0.1pt}} p{0.01pt} @{\hspace{4pt}}  >{\raggedleft}p{45pt} @{\hspace{2.5pt}} | M M M | M M M} 
        & & \multicolumn{3}{c}{Target domain $\mathcal{T}$} & \multicolumn{3}{c}{Source domain $\mathcal{S}$} \\
        & $p_\text{sup}$ & \multicolumn{1}{c}{10\%} & \multicolumn{1}{c}{50\%} & \multicolumn{1}{c|}{90\%} & \multicolumn{1}{c}{10\%} & \multicolumn{1}{c}{50\%} & \multicolumn{1}{c}{90\%} \\
        
        & {\hspace{-10pt} \fontsize{8}{0}\selectfont Domain size} & \multicolumn{1}{c}{\fontsize{9}{0}\selectfont 4358} & \multicolumn{1}{c}{\fontsize{9}{0}\selectfont 2421} & \multicolumn{1}{c|}{\fontsize{9}{0}\selectfont 484} & \multicolumn{1}{c}{\fontsize{9}{0}\selectfont 484} & \multicolumn{1}{c}{\fontsize{9}{0}\selectfont 2421} & \multicolumn{1}{c}{\fontsize{9}{0}\selectfont 4358} \\
        
        \hline
        \multirow{5}{*}{\rotatebox[origin=c]{90}{\fontsize{9pt}{0}\selectfont \textbf{Models}}} & \textit{Random} & \textit{100} & \textit{100} & \textit{100} & \textit{100} & \textit{100} & \textit{100} \\
        & M$(\varnothing)$ & 39.6 & 26.3 & 16.9 & 6.6 & 8.68 & 10.9 \\ 
        & M$(\mathcal{V})$ & 21.0 & 11.8 & 6.9 & 0.9 & 2.3 & 3.5 \\
        & M$(\mathcal{C}_{S_H})$ & 28.6 & 15.0 & 10.7 & 3.5 & 3.9 & 4.4 \\
        & M$(\mathcal{C}_{S_H}, \mathcal{V})$ & 18.2 & 9.4 & 6.0 & 0.8 & 1.8 & 2.4 \\
        \hline
        & $\delta_\mathcal{C}$  & \textit{13.4} & \textit{20.2} & \textit{13.4} & \textit{13.8} & \textit{24.4} & \textit{31.5} 
    \end{tabularx}
\end{table}

\section{MRR and top-$k$ performances}
ZSL models are usually evaluated with recall@$k$ or MRR (mearn reciprocal rank, i.e.\ harmonic mean). However, the metrics are not optimal to evaluate our models for two reasons:
\begin{itemize}
    \item Theoretically, recent research points out that RR is not an interval scale and thus MRR should not be used (Fuhr, Some Common Mistakes In IR Evaluation, And How They Can Be Avoided. SIGIR Forum 2017 ; Ferrante et al. Are IR evaluation measures on an interval scale? ICTIR 2017).
    \item Practically, we make the size of the target domain vary (10\%, 50\%, 90\%). MRR and top-$k$ scores cannot be compared across these scenarios (e.g.\ top-5 among 100 entities is not comparable to top-5 among 1000)
\end{itemize}
Therefore, as explained in Section~\ref{mfr} we used MFR (mean first relevant): the arithmetic mean of rank numbers (linearly rescaled to have 100\% for random model and 0\% for perfect model). FR is an interval scale and thus can be averaged.

However, we report here top$k$ and MRR scores in Table~\ref{topmrr}.

\begin{table}[h]
    \centering
        \caption{\label{topmrr} Recall@$k$ ($k \in \{1, 5, 10\}$) (in percentage) and MRR scores (in percentage). $p_\text{sup}=50\%$.}
    \begin{tabularx}{\columnwidth}{@{\hspace{0.1pt}} r @{\hspace{0.1pt}} | M M M c@{\hspace{1pt}} | M M M c@{\hspace{1pt}}}
        & \multicolumn{4}{c}{Target domain $\mathcal{T}$} & \multicolumn{4}{c}{Source domain $\mathcal{S}$} \\
        & \multicolumn{3}{c}{Recall @} & \multirow{2}{*}{MRR} & \multicolumn{3}{c}{Recall @} & \multirow{2}{*}{MRR} \\
        & 1 & 5 & 10 & & 1 & 5 & 10 & \\
        \hline
        {\fontsize{9pt}{0pt}\selectfont \textit{Random}} & \textless.1 & 0.2 & 0.4 & \textless.1 & \textless.1 & 0.2 & 0.4 & \textless.1 \\
        {\fontsize{7pt}{0pt}\selectfont M$(\varnothing)$} & 3.2 & 11.7 & 16.3 & 7.8 & 5.7 & 17.9 & 24.9 & 12.5 \\ 
        {\fontsize{7pt}{0pt}\selectfont M$(\mathcal{V})$} & 14.7 & 33.5 & 43.2 & 24.0 & 36.3 & 63.8 & 73.1 & 48.8 \\
        {\fontsize{7pt}{0pt}\selectfont M$(\mathcal{C}_{S_H})$} & 5.9 & 17.8 & 25.4 & 11.9 & 17.3 & 43.7 & 56.7 & 29.9 \\
        {\fontsize{7pt}{0pt}\selectfont M$(\mathcal{C}_{S_H}, \mathcal{V})$} & 15.0 & 34.7 & 44.7 & 24.7 & 41.6 & 70.6 & 78.6 & 54.2
    \end{tabularx}
\end{table}

\end{document}